# ALIGN: A Vision-Language Framework for High-Accuracy Accident Location Inference through Geo-Spatial Neural Reasoning

MD Thamed Bin Zaman Chowdhury[a*],
Moazzem Hossain[a]

[a]*Department of Civil Engineering,*
*Bangladesh University of Engineering and Technology,*
*Dhaka-1000, Bangladesh*

## ABSTRACT

In low- and middle-income countries, public safety and urban planning initiatives frequently face a critical shortage of accurate, location-specific road crash data. Extracting reliable geospatial information from unstructured text requires overcoming the limitations of traditional text-based geocoding tools, which often fail in multilingual environments with ambiguous place descriptions. This study introduces ALIGN (Accident Location Inference through Geo-Spatial Neural Reasoning), a vision-language framework designed to emulate human spatial reasoning to infer precise accident coordinates from unstructured Bangla news reports and map-based cues. A multi-stage automated pipeline was developed to process diverse textual and visual data, integrating large language models for cue extraction with vision-language models for map verification. Using an agentic architecture, we modelled an iterative reasoning loop that combines Optical Character Recognition (OCR), grid-based spatial scanning, and a 3-run geometric voting method to mathematically isolate and reduce visual hallucinations. The findings highlight that the multimodal ALIGN framework significantly outperforms traditional text-only geoparsing baselines. For example, the proposed system successfully reduced the mean localization error from an unusable 10.915 km to a sub-kilometer precision of 0.593 km on a validation dataset. Furthermore, testing the framework against official Dhaka Metropolitan Police records confirmed its reliability by achieving a mean error of 0.465 km. The results provide a high-accuracy, training-free foundation for automated crash mapping in data-scarce regions, supporting evidence-driven road-safety policymaking and the integration of multimodal AI in transportation analytics.

* Corresponding author
E-mail addresses: zamanthamed@gmail.com (MTBZ Chowdhury), moazzem@ce.buet.ac.bd (M. Hossain)

## 1. Introduction

Accurate, fine-grained geospatial data is the bedrock of effective urban transport planning, mobility research, and targeted spatial safety policy. Within the domain of road safety, knowing the precise geographic location of traffic crashes is essential for diagnosing high-risk spatial clusters, deploying emergency services, and evaluating the network-wide impact of engineering interventions. The necessity for such high-resolution spatial data is underscored by the evolution of road safety geographic research, which relies heavily on precise crash coordinates to enable robust network kernel density estimation, spatial clustering, and hotzone identification (Steenberghen et al., 2010; Xie & Yan, 2013; Young & Park, 2014). While high-income nations increasingly rely on robust, integrated crash databases and vehicle telematics (Guo et al., 2022; Szpytko & Nasan Agha, 2020), utilizing advanced spatial methods such as connected-vehicle data for urban near-crash analysis (Li et al., 2026) and micro-level zonal crash prediction models (Huang et al., 2016), a significant 'geospatial data desert' persists in most Low- and Middle-Income Countries (LMICs) (Chang et al., 2020; Mitra & Bhalla, 2023). While advanced spatial models now utilize street view imagery and computer vision to identify micro-level pedestrian and non-motorist crash clusters (Miao et al., 2025), these methods remain fundamentally inapplicable in regions where even foundational location data is missing. This gap is particularly acute given that these regions bear the overwhelming brunt of global road traffic fatalities—a data void that severely exacerbates spatial inequity in urban safety and infrastructure planning (Yuan & Wang, 2021).

To examine this challenge, this research focuses on Bangladesh, a low-resource nation that exemplifies the geospatial data scarcity pervasive in the Global South. The economic toll is staggering, with traffic crashes costing up to 5.1% of the country's Gross Domestic Product (World Bank, 2022). More critically, the official infrastructure for collecting and mapping road crash data is severely underdeveloped, rendering spatial analysis and evidence-based policymaking nearly impossible. The World Bank (2022) characterizes the current system of recording, analyzing, and reporting crashes in Bangladesh as cumbersome, error-prone, and unsuitable for spatial benchmarking. Weak coordination and inadequate organizational capacity result in profound underreporting; in fact, there is an estimated 90 percent discrepancy between recorded fatalities and World Health Organization estimates, with critical crash data from 2016 onward largely absent (World Bank, 2022).

In this spatial data vacuum, unstructured public text—particularly online news articles—remains one of the few viable, real-time sources for incident-level geographic information. However, translating narrative descriptions into precise, coordinate-level locations presents a formidable technical challenge (Bigham et al., 2009). Traditional automated systems, known as geoparsers, typically rely on a text-only pipeline: extracting location names via Named-Entity Recognition (NER) and querying a geocoding API to retrieve coordinates (Gritta et al., 2018; Miler et al., 2016). This approach is inherently brittle and frequently fails in linguistically complex, low-resource environments like Bangladesh, where spatial descriptions are laden with colloquial place names, ambiguous landmarks, and non-standardized road networks. As our baseline experiments confirm, this conventional method yields a mean localization error of 10.915 km, rendering it practically useless for fine-grained spatial safety analysis.

The fundamental flaw in existing geoparsing frameworks is their absence of true geospatial reasoning. When a human geographer or spatial analyst is tasked with inferring a location, they do not merely geocode an isolated term. Instead, they synthesize the narrative to form a spatial hypothesis, consult a digital map, and visually verify whether the cartographic evidence aligns with the textual context (e.g., "Is this specific landmark situated near the mentioned highway?").

If the location remains ambiguous, the expert intuitively conducts a systematic, grid-like scan of the surrounding geographic area to visually pinpoint the site.

Driven by these methodological limitations and the urgent need for spatial data equity, this study asks: (i) Can agentic vision-language reasoning recover crash coordinates from complex, unstructured Bangla news narratives at a precision sufficient for micro-level hotzone analysis? (ii) Does the resulting high-fidelity geocoded corpus demonstrate that a nationally representative, sub-kilometer spatial dataset can be successfully synthesized from the 'data desert' of unstructured LMIC news reports?

To answer these questions, this study introduces ALIGN (Accident Location Inference through Geo-Spatial Neural Reasoning), an agentic vision-language framework designed to automate this human-expert multimodal verification process. To the best of our knowledge, ALIGN represents the first text-to-geolocation framework to integrate Vision-Language Models (VLMs) and Optical Character Recognition (OCR) to perform iterative, map-based spatial reasoning on unstructured accident reports.

The main contributions of this paper can be summarized as follows:

- The design and implementation of a multi-stage multimodal GeoAI framework that mimics human spatial reasoning by integrating LLM-based linguistic extraction with VLM-driven visual map verification.
- A novel visual-based pipeline that achieves state-of-the-art, sub-kilometer accuracy in text-to-geolocation tasks within highly challenging, data-sparse environments, successfully reducing baseline mean localization error to sub-kilometer precision.
- The introduction of a novel geometric voting method: a 3-run spatial self-consistency loop designed specifically to mathematically isolate and reduce the severe geospatial hallucinations inherent in visual-language models.
- A demonstration of geographic capability, proving that a reliable, sub-kilometer spatial dataset can be successfully synthesized across diverse national regions and high-density urban environments, establishing a foundation for quantifiable hotzone analysis in data-scarce geographic contexts.

## 2. Literature Review

This section reviews key studies addressing the extraction of geographic coordinates from unstructured text, focusing on the evolution of text-to-location inference, domain-specific geoparsing, and the emergence of AI-driven multimodal reasoning. Beyond standard geoparsing, natural language processing applied to user-generated content and social media has been successfully utilized to extract spatial accessibility issues and latent mobility risks (Wang, Y., 2026).

Systems like CLAVIN and OpenSextant established foundations using context-aware filters to distinguish place names (James, 2020; OpenSextant, 2024), while Mordecai introduced contextual embeddings to resolve ambiguous toponyms (Halterman, 2017). To address informal text, the LNEx framework employed n-gram models to handle abbreviations in social media, significantly outperforming standard NER (Al-Olimat et al., 2018). However, these early systems lacked spatial or visual reasoning.

Efforts to adapt geoparsing to safety-critical domains followed. Idakwo et al. (2025) developed a traffic crash geoparser for Nigeria using spaCy and Google's API, while Ajanaku (2025) integrated DistilBERT and BERT models for broader disaster monitoring. Although achieving high textual extraction accuracy, these frameworks relied on conventional geocoding. Shifting toward distance-based evaluation, Milusheva et al. (2021) utilized a semi-automated pipeline for Nairobi, defining accuracy within a 500 m radius. Similarly, Algiriyage et al. (2022) presented DEES, leveraging OpenStreetMap to detect disaster events in real-time without proprietary APIs. Recent developments in transport geography have increasingly mobilized large-scale, data-driven analytics to map the dynamic vulnerability of urban transit networks (Pan et al., 2024), further highlighting the growing reliance on high-fidelity spatial features.

Recent advances in Large Language Models (LLMs) have introduced new paradigms. Ling et al. (2026) emphasize the role of LLMs in constructing transportation knowledge graphs. Hu et al. (2023) utilized prompt engineering to teach GPT-4 implicit spatial relations. Furthermore, Wu et al. (2025) proposed GeoSG, a self-supervised framework bridging linguistic and spatial reasoning. Concurrently, the use of large language-and-vision models is emerging within transport geography, such as leveraging multimodal AI on satellite imagery for low-resource infrastructure assessment (Wang et al., 2026). Despite these advances, a critical gap remains for crash geolocation: no existing system fuses visual map understanding with textual extraction to perform perceptual verification. This need for precision is critical because traffic conflict risks exhibit significant spatial dependence, meaning that inaccuracies in initial geolocation can lead to a complete misunderstanding of spatial clustering and how risks propagate through the road network (Xie & Yan, 2013).

Across all reviewed studies, three core limitations persist:

- Textual Dependence – Current systems depend exclusively on linguistic cues and lack the ability to cross-validate through visual or spatial context.
- Ambiguity Resolution – Handling multiple places mentions or vague spatial relations remains a major bottleneck.
- Low-Resource Adaptability – Most frameworks are trained on English corpora and perform poorly in languages with irregular orthography or transliteration, such as Bangla.

The integration of Vision–Language Models (VLMs) offers a compelling solution to these gaps by combining map-level perception with textual inference. A multimodal geoparsing framework that can read map screenshots, verify labels visually, and perform grid-based spatial search—such as the proposed ALIGN system—would represent a new generation of geospatial AI capable of achieving human-expert-level accuracy in coordinate inference.

## 3. Methodology

Recent advances in Large Language Models (LLMs) such as OpenAI's GPT, Google's Gemini, and Meta's LLaMA have enabled powerful reasoning capabilities across text and vision (Bharathi Mohan et al., 2024; Gallifant et al., 2024; Gemini Team, Google, 2024; Touvron et al., 2023). These models can interpret complex queries and produce coherent, context-aware responses (Brown et al., 2020). However, traditional LLMs are limited in accessing real-time or domain-specific data sources, which restricts their applicability in dynamic, geospatial tasks. Retrieval-Augmented Generation (RAG) mitigates this limitation by integrating retrieval and generation, allowing models to query external databases or APIs during inference (Gao et al., 2024; Lewis et al., 2020). Building on this principle, ALIGN adopts an agentic architecture, where an LLM coordinates multiple subsystems—text extraction, OCR validation, and vision–language reasoning—to execute structured actions autonomously (Gebreab et al., 2024; Wang et al., 2024). This lightweight, API-driven design eliminates the need for GPU-intensive training and ensures a cost-efficient and highly accurate framework suitable for low-resource environments.

ALIGN operationalizes this human-in-the-loop logic through a novel, multi-stage pipeline. First, a Large Language Model (LLM) extractor parses unstructured Bangla news to identify structured location cues (roads, landmarks, administrative zones). The system then queries Google Maps (Google, 2025b) but, crucially, does not trust the result. It captures map screenshots, uses Optical Character Recognition (OCR) to read map labels, and deploys a Vision-Language Model (VLM) to visually confirm if the map screenshot is consistent with the article's narrative. If this first-stage search fails, a second-stage reasoning module activates, mimicking a human's spatial search by performing an automated grid scan of the likely area to find the location visually. We developed this system within the context of Bangladesh not only due to the urgent need but also because its linguistically complex and ambiguous place names provided a rigorous testbed. This choice was further guided by the availability of domain experts essential for manually creating the high-quality ground-truth dataset used for validation.

### *3.1 System Architecture*

The ALIGN pipeline is structured into four sequential stages: (1) text classification & cue extraction, (2) first-stage geospatial reasoning, (3) second-stage grid-based refinement, and (4) fail-safe fallback. The flowchart shown in Fig. 1 summarizes the interactions among these components.

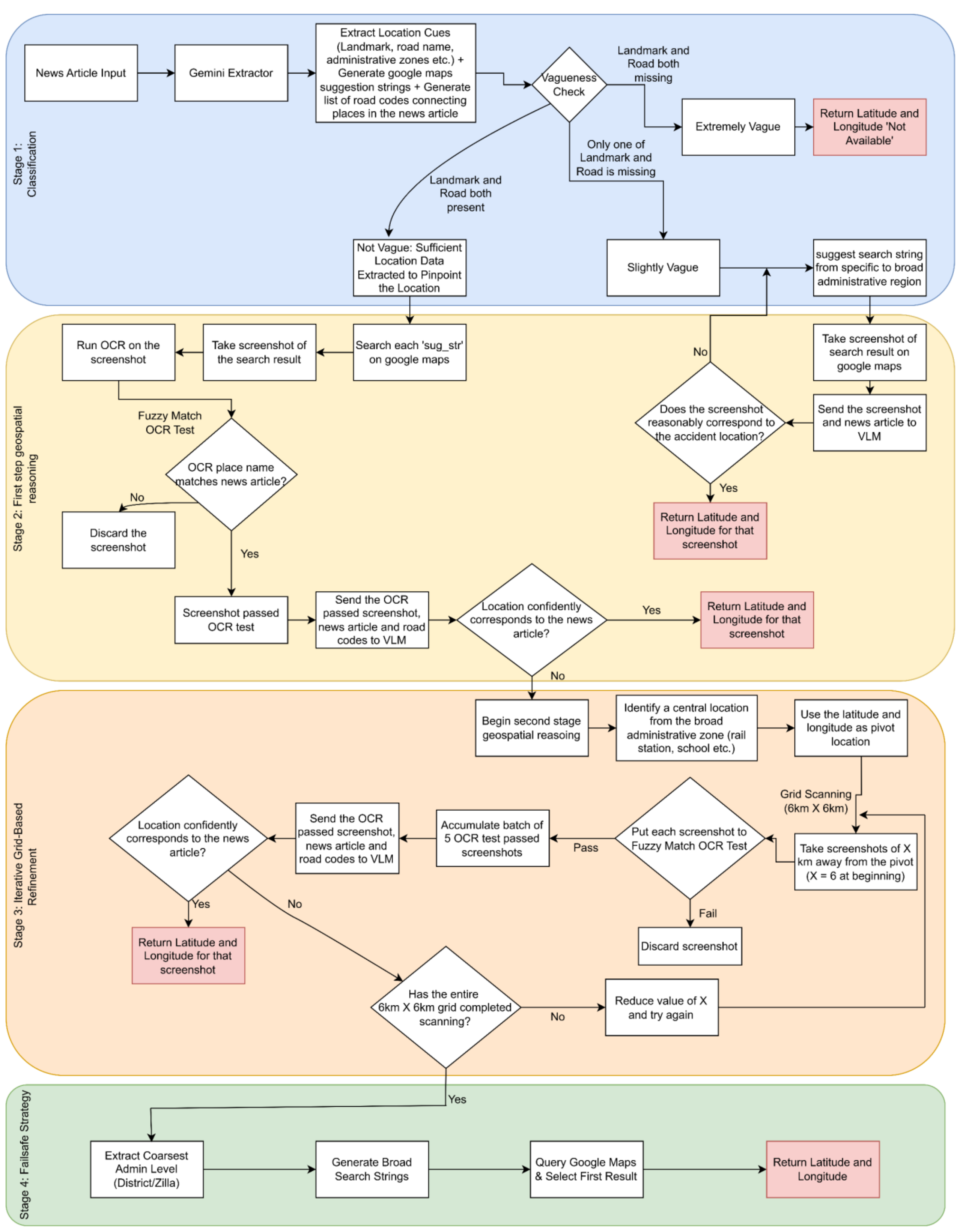

**Fig. 1.** System Architecture of ALIGN

*3.1.1 Stage 1: Text classification and cue extraction*

The pipeline initiates by feeding the unstructured Bangla news article into the Vision-Language Model (VLM). This extraction phase utilizes a highly structured, schema-guided prompt designed to systematically parse explicit location data. The model identifies and categorizes spatial mentions into a predefined hierarchical schema containing eight specific geographic fields. These include the road name, road type, and localized landmarks, alongside specific Bangladeshi administrative tiers: the union (the smallest rural administrative and local government unit), the thana (police precinct boundaries), the upazila (sub-districts), and the zila or district (the primary regional administrative level).

Based on these extracted entities, the VLM dynamically generates a comprehensive set of potential map search queries by formulating various permutations of the spatial data (for example, combining a landmark with its sub-district, or a road name with a landmark). To maximize the subsequent search retrieval hit rate, the system applies heuristic rules to these queries, generating variations that account for short or long forms, fused or segmented Bangla phrasing, cross-language transliterations, and semantic synonyms. It also creates root-token fallbacks by stripping generalized facility identifiers (such as "curve" or "petrol pump") to isolate the core geographic name. Furthermore, the model produces a bilingual inventory of these place names to facilitate cross-lingual map searches.

Prior to executing any map queries, the pipeline normalizes the extracted place names using a fuzzy alias map. This resolves common discrepancies in alternate English transliterations of Bangla locations (e.g., reconciling "Jashore" with "Jessore"). The normalized names are then cross-referenced against a structured national database of Bangladeshi road networks (bd_roads.json) using fuzzy logic. If an extracted colloquial road name matches an official government road designation ("List of roads in Bangladesh," 2025), the corresponding alphanumeric route code (e.g., National Highway N1 or Regional Road R375) is appended to the context. This injection mechanism is critical to bridging the VLM's inherent spatial knowledge gaps regarding regional road numbering. By explicitly grounding the model with these official route codes in subsequent stages, the pipeline prevents spatial hallucinations and dramatically improves localization accuracy.

Finally, a deterministic vagueness check is executed as a pre-filtering guardrail. If the text lacks both a specific localized landmark and a road name, the description is classified as extremely vague. To prevent the VLM from blindly guessing, the system terminates the process and returns a "Latitude and Longitude Not Available" status. If only one of these two critical cues is missing, the case is flagged as slightly vague and routed to specialized fallback logic in the subsequent geospatial reasoning stages. Otherwise, fully detailed cases proceed directly to the multimodal visual verification pipeline.

*3.1.2 Stage 2: First-stage geospatial reasoning*

In this stage, the pipeline attempts to pinpoint the accident location by systematically querying Google Maps using the varied search strings generated during the initial extraction phase. To maximize computational efficiency and minimize API costs, this search process is structured into three highly optimized sub-steps.

The first step is automated autocomplete suggestion retrieval. Using a headless web browser with a fixed viewport, the system simulates a user typing each query into the map's search interface and captures the resulting dropdown suggestions. Utilizing headless browsing with a disabled GPU ensures environmental reproducibility while effectively preventing rate-limiting issues during high-volume queries.

The second step involves filtering and reranking these captured suggestions. The system initially applies simple rule-based filters to eliminate generic or invalid interface options (such as "Add a missing business"). The remaining viable candidates are then batched and sent to the VLM in a single call. The model is presented with the original news article alongside grouped candidate suggestions, and is tasked with selecting the most contextually relevant option from each group. To conserve computational tokens, the model is strictly instructed to output its selections without generating a reasoning trace (Google, 2025a). This reranking process guarantees that the system prioritizes highly probable, real-world locations, preventing it from wasting time on irrelevant or geographically distant suggestions.

The third step is screenshot capture and rigorous Optical Character Recognition (OCR) verification. For each highly-ranked suggestion, the automated driver navigates to the location, allows the map tile to fully render, and captures a full-window screenshot. To overcome the inherent limitations of default VLM text recognition on low-resolution Bangla text, these screenshots are processed through a custom, chunk-wise OCR agent utilizing EasyOCR (Jaided AI, 2025). The map image is divided into overlapping blocks, ensuring small, dense, or rotated map labels are not missed. Each chunk undergoes targeted image preprocessing—including grayscale conversion, binary thresholding to emphasize text, and a 2x upscale before extraction. The resulting bilingual text lines are normalized to remove punctuation and compared against the article’s required place names using a hybrid fuzzy matching algorithm from the thefuzz library (SeatGeek, 2025). If any detected map label achieves a similarity score of 75% or higher with a target place name, the screenshot passes the test; otherwise, it is immediately discarded. This deterministic OCR pre-filtering acts as a strict guardrail, drastically reducing the number of expensive VLM image evaluations by discarding visually irrelevant map tiles early in the pipeline.

Finally, all screenshots that successfully pass the OCR filter are forwarded to the VLM for visual verification. The model evaluates the original article text, the filtered map screenshot, and any injected official road codes to determine if the map's visual layout precisely corresponds to the described accident site. The VLM outputs a definitive binary decision alongside a brief justification for its reasoning. If the location is confirmed, the system extracts the exact geographic coordinates (latitude and longitude) from the active URL and successfully terminates the search. If rejected, the system seamlessly moves to the next highest-ranked suggestion. Ultimately, the integration of batch reranking and deterministic OCR pre-filtering guarantees that the VLM only expends computational resources evaluating map views that demonstrably contain the correct geographic context.

### *3.1.3 Stage 3: Second-stage grid-based refinement*

If the first stage fails to locate a match, the system enters a more exhaustive grid-based search. This situation typically arises in slightly vague cases where only one of the cues (landmark or road) is present. The pipeline first identifies the broadest confidently extracted administrative unit (e.g., union or upazilla). This becomes the pivot location. A 6 km × 6 km grid is centered on the pivot's coordinates, and the algorithm generates search queries at regularly spaced grid points (initial step size 6 km). For each grid point, the pipeline performs the same sequence as in Stage 2: search on Google Maps (entering the grid coordinates directly), capture a screenshot, run the OCR test, and send the screenshot through Prompt 2. If no match is found after scanning the entire grid, the step size is halved (6 km → 3 km → 1 km) and the process repeats. This recursive scanning continues until a match is found or the finest grid spacing is exhausted. Only screenshots that mention the accident's road or landmark via OCR pass on to the VLM. The grid search is computationally intensive but ensures coverage when explicit cues are insufficient.

### *3.1.4 Stage 4: Fail-safe fallback*

In rare cases where both the first and second stages fail; the system falls back to a Slightly Vague strategy. The pipeline reverts to the coarsest administrative level (district) and generates broad search strings, gradually narrowing down the region. This fallback ensures that the system always returns the best available approximation, even when the description is extremely vague or incomplete.

## *3.2 Techniques in Logical Stages*

This section elaborates on the techniques used in each stage, highlighting the rational decisions that underpin ALIGN's performance.

### *3.2.1 Text cue extraction*

**Reasoning motive.** Bangla news articles often describe accidents using a mix of colloquial place names, abbreviations and administrative units. Traditional rule-based extraction fails to capture the breadth of variations. We therefore leverage a large language model to parse the article. Prompt 1 instructs the model to emit structured fields and to generate a curated set of Google Maps search strings. These search strings are central to the subsequent search, as we observed in our preliminary testing that the relevance of Google Maps results depended heavily on the exact phrasing of the queries. The cross-language expansion rules in the prompt address transliteration issues between Bangla and English by generating both scripts and by creating synonyms for facility types (e.g., "mosque"). The alias map corrects alternative spellings (e.g., "chattogram"→"chittagong") and normalises numbers in road names. Without this step, the system would often search for nonexistent variations and fail to find the accident site.

**Vagueness detection.** The vagueness check uses simple heuristics: if the article lacks a landmark and a road, the case is extremely vague. If exactly one cue is missing, the article is labelled slightly vague and triggers Stage 3 if the first search fails. This triaging prevents wasteful grid searches when precise cues exist and ensures resilience when they do not.

**Road code injection.** A unique nuance of our pipeline is the use of a road code lookup. Our preliminary testing indicated that the VLM's knowledge of Bangladeshi roads is incomplete; for instance, we observed that a regional highway might be known by its colloquial name (e.g., "Sunamganj-Sylhet Highway") but not by its official code (e.g., Z260). By fuzzy matching extracted place names against an external road database, we obtain the relevant road code and pass it into subsequent prompts. This additional context allows the VLM to cross-reference the road's route with the map screenshot and dramatically reduces misinterpretations.

*3.2.2 Suggestion retrieval and reranking*

**Headless Selenium automation.** We interact with Google Maps programmatically using a Selenium WebDriver (The Selenium Project, 2025). Running in headless mode eliminates overhead from rendering the user interface while preserving the exact DOM structure necessary for capturing suggestions and map screenshots. We fix the viewport at 1920×1080 pixels to ensure consistent screenshot dimensions; this matters because OCR performance is sensitive to text size and resolution. The WebDriver disables GPU acceleration and uses a large shared memory to avoid crashes when loading heavy pages.

**Autocomplete filtering.** Google Maps often includes generic prompts in its suggestions (e.g., "Add a missing business or landmark"). These are removed by keyword matching before reranking. Filtering ensures that the VLM does not waste time evaluating non-location suggestions.

**Reranker (Prompt 4).** Instead of selecting suggestions by simple heuristics (e.g., first suggestion), we employ a VLM reranker to choose the most contextually relevant suggestion. Prompt 4 presents the news article and grouped suggestions and instructs the VLM to select exactly one suggestion per group. This single call returns the best suggestions for all search strings, reducing the number of VLM calls compared with per-query reranking. The design of the prompt emphasizes conciseness: the model returns only the <best_x> tags without any explanatory text, minimizing token usage. This design choice, coupled with batching, contributes to both efficiency and cost reduction (fewer API calls and less output).

*3.2.3 Chunk-wise OCR and fuzzy matching*

**Motivation for custom OCR.** Google Maps embeds place names as part of the map canvas rather than in accessible HTML elements. The built-in OCR capabilities of Gemini or other VLMs struggle with Bangla text due to font variations and small sizes. We therefore integrate EasyOCR, a deep learning–based OCR library that supports Bangla (bn) and English (en) scripts. To avoid missing small or rotated labels, the screenshot is subdivided into overlapping chunks (default 800 × 400 pixels with 50 pixels overlap). Each chunk is pre-processed by converting to grayscale, applying an inverted binary threshold (to highlight white text on dark backgrounds) and upscaling by 2× to enlarge characters. Running OCR on each chunk separately yields more detected text lines than processing the whole image at once.

**Hybrid fuzzy matching.** The OCR agent cleans all text lines by normalising them (NFC normalisation, removal of non-alphanumeric characters and extra spaces) and discards lines shorter than three characters. It then compares each place name extracted from the article to the OCR lines using a hybrid similarity score: the average of token_sort_ratio and partial_ratio from the thefuzz library. token_sort_ratio rearranges tokens to ignore word order, while partial_ratio matches substrings; their combination captures both global and local similarities. A threshold of 75% was chosen empirically: it balances false positives (random map labels) and false negatives (genuine but slightly misspelled names). If any match exceeds the threshold, the screenshot is considered relevant and passed to the VLM for final judgment. Batching and fuzzy matching together reduce the number of images requiring expensive multimodal reasoning.

#### *3.2.4 VLM-based screenshot verification and grid search*

**Prompt 2 (Screenshot Verifier).** The OCR-approved screenshots are passed to a multimodal prompt. This prompt includes the news article, the screenshot image, the OCR-extracted text and the road codes. The VLM is asked to assess whether the image shows the same location as the accident described in the article, returning <isSame> Yes/No <isSame> and a short justification. Providing OCR text and road codes helps the model ground its reasoning; without them, the VLM might misinterpret general map features. The prompt also instructs the model to remain concise, reducing token usage.

**Iterative grid search (Stage 3).** When the first stage fails to find a match, the pipeline resorts to an iterative grid search. The pivot location is selected from the highest-confidence administrative unit in the article. The system generates a grid of points (initial spacing 6 km) covering a 6 km × 6 km area. Each grid point is searched on Google Maps, and the same screenshot-OCR-VLM verification loop is executed. If no match is found, the grid spacing is reduced (6 → 3 → 1 km) and the search continues. This scheme ensures complete coverage while controlling computational cost: the step size reduction is only invoked if previous passes fail.

**Fail-safe fallback (Stage 4).** If all grid searches fail, the system falls back to a simple strategy: it escalates search strings from specific to broad administrative levels (e.g., from union to district) and returns the first coordinates found. Although coarse, this fallback prevents the system from returning null results in extremely vague cases.

### *3.3 Vision-Language Models and Implementation Parameters*

To ensure the ALIGN framework's reproducibility and to establish its generalizability across different foundation models, the system architecture was benchmarked using three distinct Large Language Models (LLMs) and Vision-Language Models (VLMs). Rather than relying on a single proprietary engine, we evaluated Gemini 2.5 Flash (Gemini Team, 2024; CodeGPT, 2025), GPT-5-mini (OpenAI, 2026), and Llama-4-Maverick (FP8) (Meta AI, 2026; Together AI, 2026) . Model parameters were strictly controlled, heavily favoring native provider defaults to minimize hyperparameter variance, except when enforcing structured JSON outputs. Concurrently, the deterministic map-label verification module was standardized using specific chunk-wise EasyOCR preprocessing configurations. The exact initialization parameters, model specifications, and verification constraints utilized in this study are detailed in Table 1.

**Table 1**
VLM and OCR parameters

| System Component | Provider (Date/Version) | Parameter Category | Exact Settings & Constraints |
|---|---|---|---|
| Gemini 2.5 Flash | Google (Dec 2025) | Generative Limits | Temperature: 1.0; Top-p: 0.95 |
| | | Token Budget | Max output limit: 8,192 tokens |
| | | Utilization | Primary multimodal reasoning baseline |
| GPT-5-mini | OpenAI (Jan 2026) | Generative Limits | Temp & Top-p explicitly disabled (internal reasoning) |
| | | Token Budget | Constrained strictly via max_completion_tokens |
| Llama-4-Maverick (17B FP8 Instruct) | Together AI (Feb 2026) | Generative Limits | Temperature: 0.7; Top-p: 0.7 |
| | | Token Budget | Max output limit: 16,384 tokens |
| | | Output Strategy | Enforced strict reliability via JSON schema modes |
| EasyOCR Pipeline | Jaided AI (v1.7.2) | Initialization | Languages: Bengali ('bn'), English ('en'); GPU Enabled |
| | | Tile Chunking | 800x400 pixels with a 50-pixel overlap window |
| | | Pre-processing | Grayscale, 2x cubic upscaling, inverse binarize (170) |
| | | Text Matching | NFC Unicode normalized; 75% hybrid fuzzy threshold |
| | | Local Hardware | Intel Core i9-11900H; NVIDIA RTX 3060 (6GB) GPU. |

The parameters outlined in Table 1 are foundational to ALIGN’s operational mechanics. For the reasoning engines, generative hyperparameter tuning was intentionally constrained. Gemini 2.5 Flash and Llama-4-Maverick utilized standard temperature settings to balance deterministic extraction with linguistic flexibility, while Llama-4 specifically required enforced JSON schema modes to guarantee structural reliability during extraction and reranking stages. Conversely, GPT-5-mini operates with internal reasoning pathways where standard temperature and top-p parameters are explicitly disabled, requiring constraint management purely through completion token budgets. Beyond the generative models, the EasyOCR configurations serve as a critical deterministic gatekeeper as shown later in the ablation study.

### *3.4 System Realization: The Sunamganj Bus Accident Case*

This section illustrates how ALIGN logics and mechanisms successfully localized an accident site described in a Bangla news article despite incomplete map data, showcasing its two-step geospatial reasoning strategy.

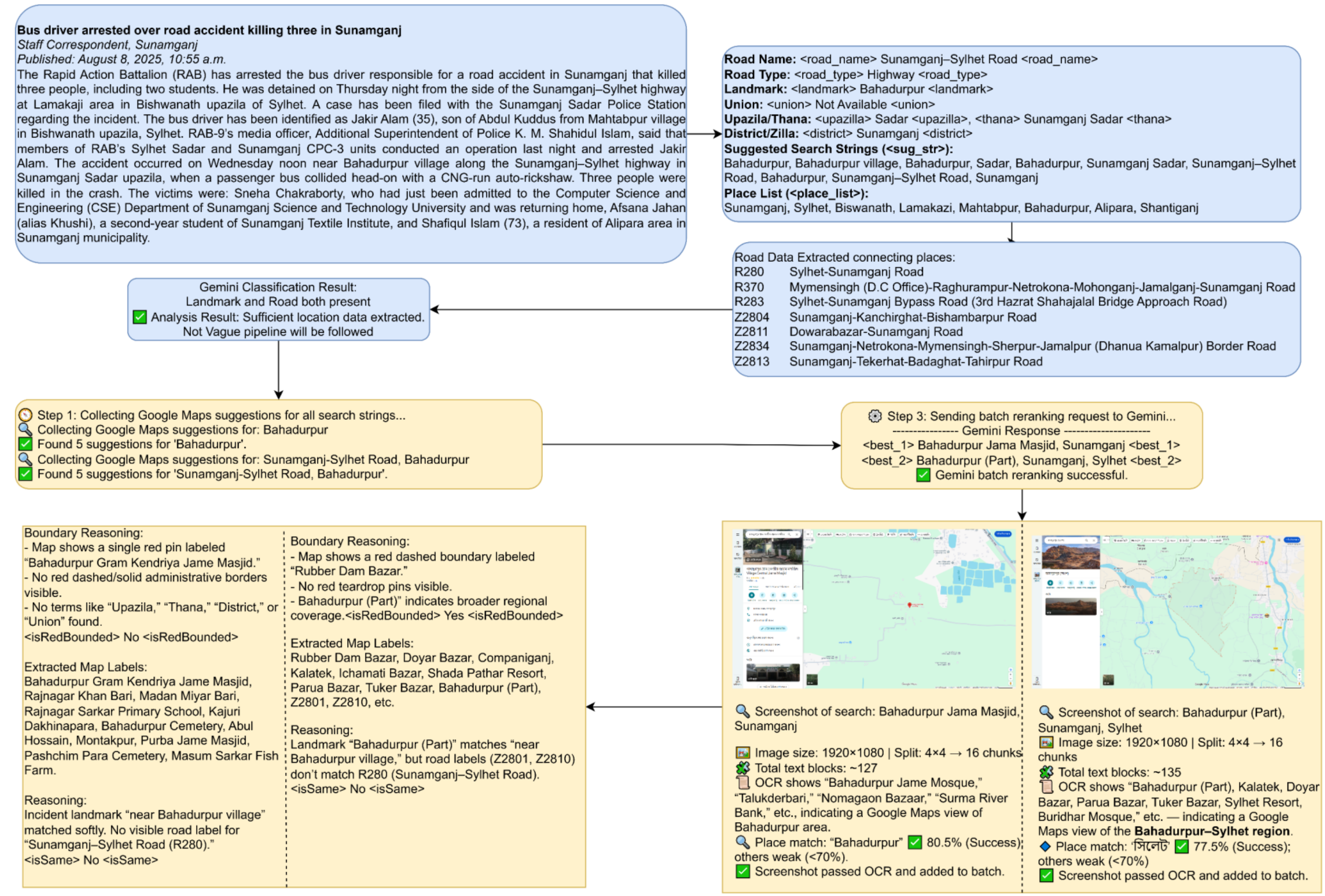

**Fig. 2.** First-Stage Reasoning (Initial Extraction & Search Failure)

Fig. 2 illustrates the initial extraction and first-stage reasoning of the ALIGN framework using a sample news report about a bus accident. The pipeline begins by extracting structured cues—specifically the "Sunamganj–Sylhet Road," the landmark "Bahadurpur," and the administrative district—classifying the case as "Not Vague." It then generates and reranks Google Maps search strings to retrieve candidate locations. The core of this stage is the visual verification loop, where the system analyzes map screenshots using OCR and VLM reasoning. The figure displays specific attempts where the system rejects high-scoring text matches because the visual context is inconsistent with the accident description. For example, one candidate is rejected because it depicts a demarcated administrative boundary rather than a highway, while another shows an urban commercial outlet. Because the specific "Bahadurpur village" is not indexed near the highway in the standard geocoding database, the first-stage search fails to find a confirmed match, triggering the need for the advanced grid-scanning phase.

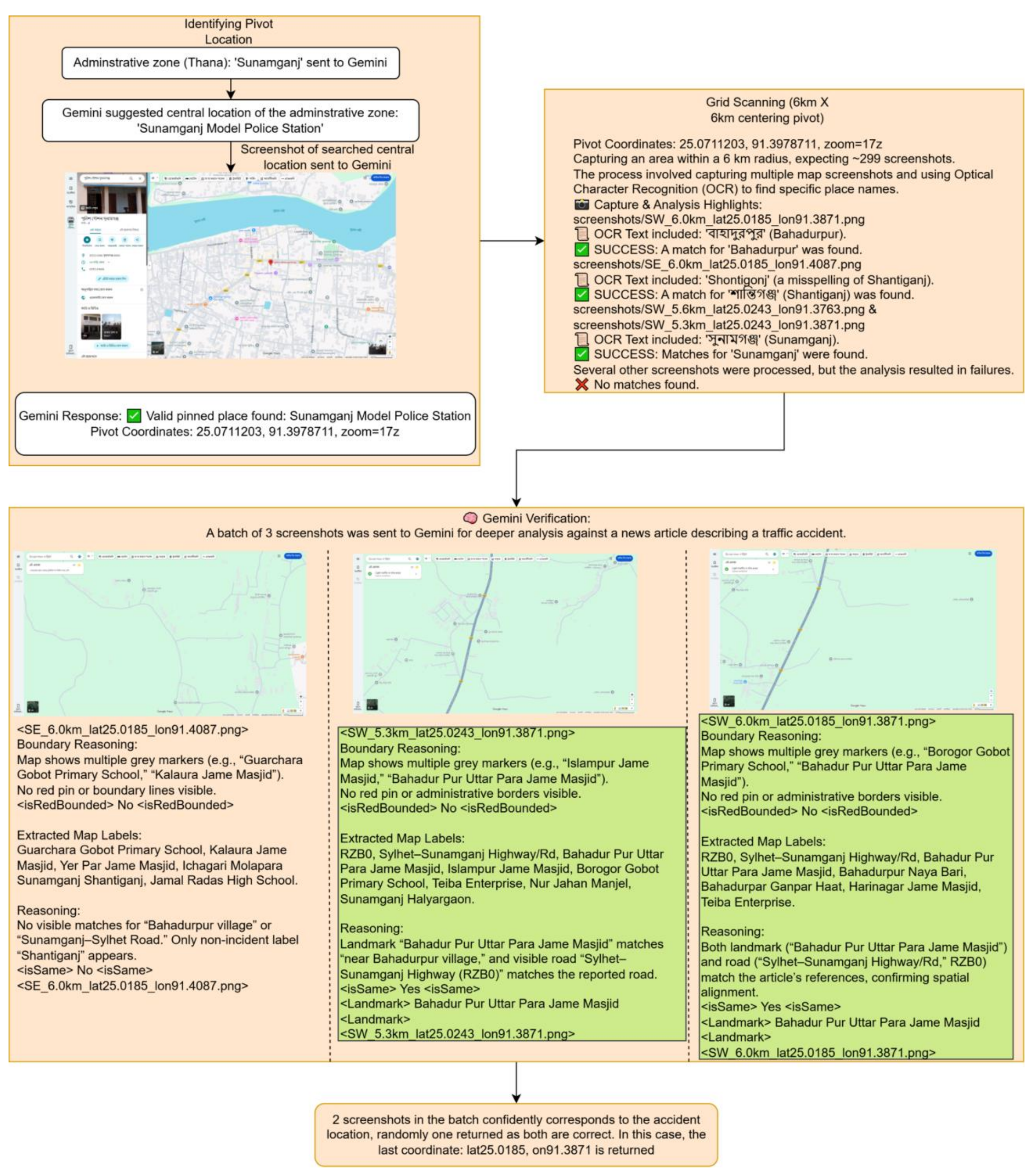

**Fig. 3.** Second-Stage Reasoning (Grid Scanning and Visual Verification)

Fig. 3 depicts the "Second-Stage Reasoning," triggered after the initial search failure. The system calculates a pivot point—the "Sunamganj Model Police Station"—and initiates a 6 km × 6 km grid scan around it to find the unindexed location visually. The process involves capturing map screenshots at regular intervals, which are filtered by OCR for relevant keywords like "Bahadurpur." The figure highlights the VLM's analysis of a batch of three filtered screenshots.

While one image is discarded for lacking relevant features, the model successfully identifies two screenshots showing "Bahadurpur Pur Uttar Para Jame Masjid" situated directly along the "Sylhet–Sunamganj Highway (R280)." Although the exact village name from the text was missing in the digital map database, the VLM correlates the visible mosque and highway labels with the accident narrative to confirm the location. The system validates this visual evidence, returning the precise coordinates (25.0185, 91.3871) and demonstrating how spatial neural reasoning resolves ambiguity where text-only geocoding failed.

### *3.5 Model Evaluation*

#### *3.5.1 Manual verification and dataset construction*

To establish ground truth, we manually geolocated each accident in a dataset of Bangla news articles. Two independent annotators read each article and used Google Maps to pinpoint the exact crash site based on contextual clues (landmarks, road intersections, nearby institutions). To objectively validate the reliability of this human baseline, an Inter-Rater Reliability (IRR) analysis was conducted. Disagreements were resolved by discussion. The dataset was split into a system development set of 36 articles (used for prompt tuning, OCR threshold selection and parameter tuning) and a validation set of 77 articles (unseen during development). Performance metrics reported in Section 4.1 are computed on the validation set.

#### *3.5.2 External Verification*

To validate the system's performance against authoritative data, we collaborated with the Dhaka Metropolitan Police (DMP). While obtaining comprehensive ground truth data for the entire country is challenging due to the limitations in official recording systems, the DMP tracks Road Traffic Crash (RTC) locations within the metropolitan area to a limited extent. For this external verification, the DMP provided the official ground-truth coordinates for a set of specific accident incidents, along with the corresponding newspaper articles covering those exact events. We then ran the ALIGN system over these matched articles to measure its performance against independent, official police records, providing a robust counter-verification to the manually annotated datasets described above.

#### *3.5.3 Metrics*

Geolocation accuracy is assessed by comparing the predicted coordinates $(\hat{\varphi}, \hat{\lambda})$ to the ground-truth coordinates $(\varphi, \lambda)$ using the Haversine distance. For two points $A = (\varphi_1, \lambda_1)$ and $B = (\varphi_2, \lambda_2)$, the great-circle distance is

$$d_{AB} = 2R \ \arcsin\left(\sqrt{\sin^2\left(\frac{\phi_2 - \phi_1}{2}\right) + \cos(\phi_1)\cos(\phi_2)\sin^2\left(\frac{\lambda_2 - \lambda_1}{2}\right)}\right) \tag{1}$$

where $R = 6{,}371$ km is the mean radius of the Earth. The resulting $d_{AB}$ represents the shortest path (great-circle distance) between the two points, making it well-suited for evaluating coordinate-level geolocation errors.

**Mean Absolute Error (MAE).** The MAE measures the average magnitude of localization errors, ignoring their direction. It provides a direct and interpretable indication of the system's overall accuracy:

$$\text{MAE} = \frac{1}{N}\sum_{i=1}^{N} \mid d_i \mid \quad (2)$$

A lower MAE value indicates that most predicted coordinates are close to their true locations.

**Root Mean Square Error (RMSE).** The RMSE emphasizes larger deviations by squaring the errors before averaging. It is particularly useful when identifying occasional high-error cases (outliers) that the model might produce:

$$\text{RMSE} = \sqrt{\frac{1}{N}\sum_{i=1}^{N} d_i^2} \quad (3)$$

Because of the squaring operation, RMSE values are always greater than or equal to MAE, highlighting how well the model controls extreme errors.

**Median Error.** The median error represents the middle value in the sorted list of all distance errors, offering a robust measure of "typical" performance that is not influenced by outliers:

$$Median_{Error} = median(d_1, d_2, \dots, d_N) \quad (4)$$

This metric reflects the distance within which 50% of all predictions fall and is useful for summarizing central accuracy in skewed distributions.

**Cumulative Distribution Function (CDF).** The CDF illustrates how errors are distributed across varying distance thresholds. For thresholds $\delta = 0.5, 1, 2,$ and 5 km, the CDF value $P(D \leq \delta)$indicates the proportion of articles whose localization error does not exceed $\delta$:

$$P(D \leq \delta) = \frac{1}{N}\sum_{i=1}^{N}[d_i \leq \delta] \quad (5)$$

This curve provides an intuitive view of model reliability — for example, a CDF (1 km) means most predicted locations are within 1 km of their actual positions.

**Inter-Rater Reliability (IRR).** While metrics such as Cohen's Kappa are traditionally used for categorical agreement, they are mathematically unsuitable for evaluating continuous spatial coordinates. Therefore, the reliability of the human annotations was quantified using the Intraclass Correlation Coefficient (ICC) for absolute agreement, alongside the Mean Spatial Deviation calculated via the Haversine formula. To measure the statistical consistency of the annotations, a two-way random effects, absolute agreement, single rater model, ICC(2,1), was applied:

$$ICC(2,1) = \frac{MS_R - MS_E}{MS_R + (k-1)MS_E + \left(\frac{k}{n}\right)(MS_C - MS_E)} \quad (7)$$

Here, $MS_R$ is the mean square for rows (targets), $MS_E$ is the mean square error, $MS_C$ is the mean square for columns (raters), k is the number of raters, and n is the number of targets.

Together, these metrics offer a comprehensive evaluation of geolocation performance. While MAE and RMSE summarize overall error magnitude, the median highlight central tendencies, and the CDF captures spatial consistency across error thresholds. Finally, ICC measures the reliability of the human annotators.

Beyond numerical metrics, we perform qualitative analysis. Annotators examine failure cases to identify patterns: misclassified landmarks, ambiguous road names or missing OCR matches. We also evaluate efficiency by counting the number of VLM calls per article and the number of grid points scanned. These statistics inform potential improvements in prompt design and search heuristics.

### *3.6 Hallucination Analysis and Prevention*

When Vision-Language Models (VLMs) are deployed for spatial reasoning in data-sparse and low-resource environments, they often suffer from severe, confidently incorrect localizations. An analysis of the system logs and extreme spatial outliers observed during the ablation study (Section 4.3) reveals that these errors are not random, but stem from specific multimodal failure modes. To address this, the ALIGN framework utilizes a comprehensive classification of these spatial hallucinations, followed by a multi-stage, defense-in-depth mitigation strategy.

#### *3.6.1 Types of VLM Spatial Hallucinations*

By examining the raw reasoning traces of the VLM across failed validation cases, we categorized the spatial hallucinations into three primary types:

**Type A: Context-Ignorant False Positives (Spatial Oversimplification)** This occurs when the VLM correctly reads the map canvas but forces a match by ignoring municipal scale or highly specific pinpoint details. The model settles for a partial, high-level geographical match rather than searching for the exact landmark.

**Type B: Multimodal Vision/OCR Hallucination** This is a critical failure where the VLM invents map labels that do not visually exist on the provided map canvas, projecting its own linguistic assumptions onto the image.

**Type C: Spatial Knowledge Gap (Road Network Misalignment)** This failure mode occurs when the VLM lacks the domain-specific geographic knowledge required to bridge the gap between colloquial descriptions in the unstructured text and official alphanumeric designations on the map. Without external structural grounding, the model fails to establish equivalence, forcing it to either blindly guess a location or falsely reject a correct one.

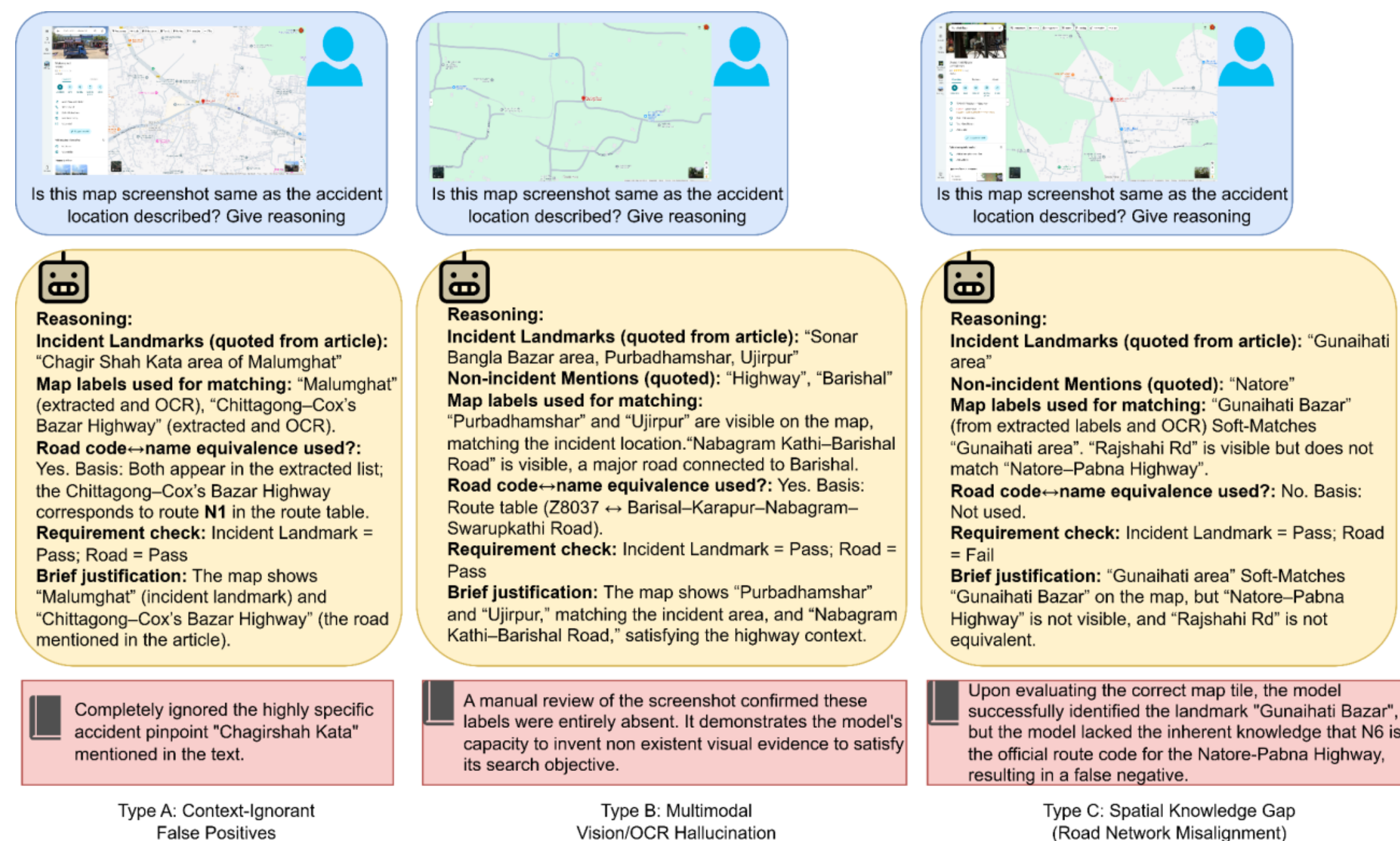

**Fig. 4.** Examples of Hallucination Types

Fig. 4 shows how the VLM hallucinates upon receiving screenshots and prompts from the user. The methods to prevent them is discussed in the next section.

*3.6.2 Hallucination Prevention Strategies*

To counteract these inherent VLM vulnerabilities, ALIGN does not rely on a single verification step. Instead, it employs a defense-in-depth architecture that actively filters hallucinations at four distinct stages of the pipeline:

**Contextual Grounding (Road Network Injection)** To prevent the model from blindly guessing the locations of colloquial highway names, official road network codes are injected directly into the prompt.

**Deterministic Pre-filtering (Chunk-wise OCR)** To directly combat Type B hallucinations (inventing visual text), the pipeline utilizes EasyOCR as a deterministic gatekeeper. Before the VLM is permitted to evaluate a map tile, the OCR module must mathematically prove the existence of the text via fuzzy matching. If the text is absent, the screenshot is discarded, physically cutting off the VLM's ability to hallucinate the label.

**Constraint via Prompt Engineering** To reduce Type C overconfidence, the system utilizes highly structured JSON schemas and explicit vagueness-check routing. The prompt forces the LLM to categorize the specificity of its findings, effectively mandating that the model admit when it lacks sufficient pinpoint data rather than guessing.

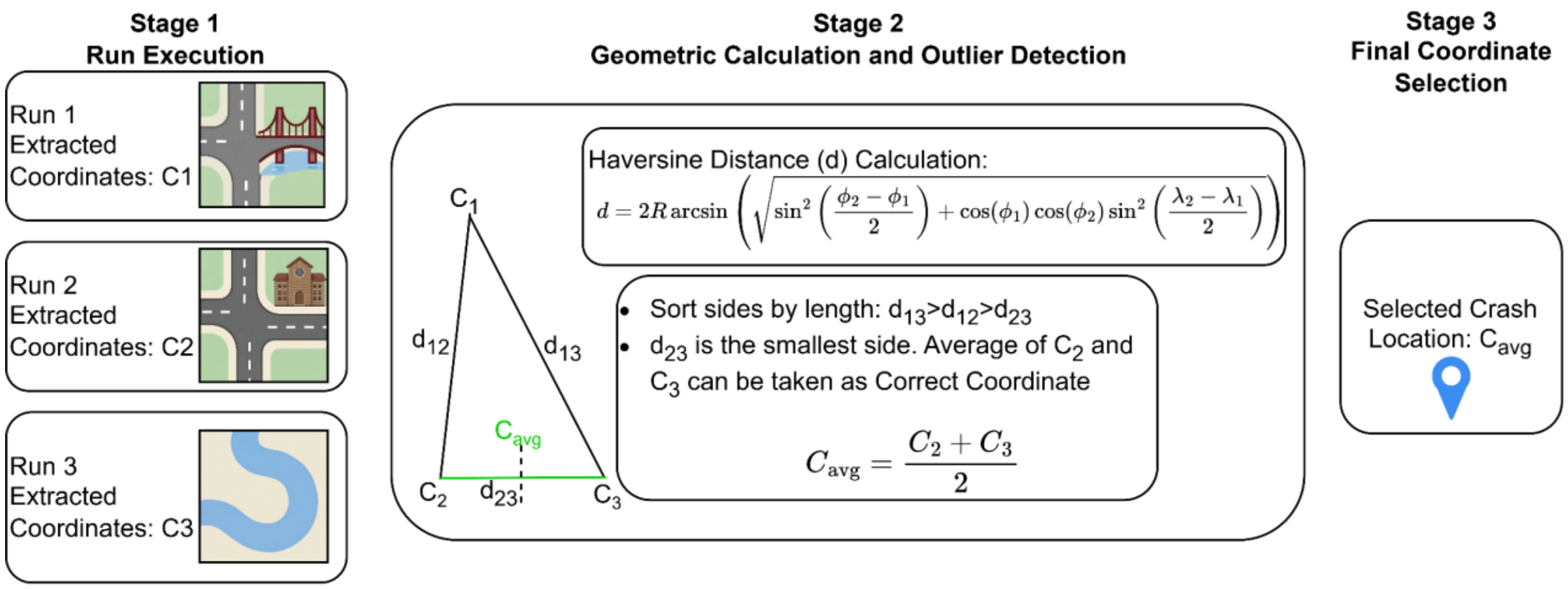

**Fig. 5.** Geometric Voting (Three Runs)

**Geometric Voting** For the residual hallucinations that manage to bypass the OCR and injection filters, ALIGN deploys a mathematical fail-safe: a 3-run spatial self-consistency loop (Wang et al., 2022). The system assumes that while a model may hallucinate once, it is statistically unlikely to produce the exact same severe spatial hallucination across multiple independent runs.

The method calculates the Haversine distance between coordinates generated from three independent runs. The process is illustrated in Fig. 5.

## 4. Results and Discussions

### *4.1 Quantitative Performance Evaluation*

This section evaluates ALIGN's geolocation accuracy across three distinct datasets: a system development set, an unseen validation set for generalizability, and an external verification set from the Dhaka Metropolitan Police. These comparisons establish the system's precision and reliability against both manual annotations and official ground-truth records.

However, before evaluating the framework's predictive performance, it is necessary to establish the validity of the manually annotated ground-truth coordinates. As detailed in Section 3.5.1, two independent annotators geolocated the accident sites based on the textual narratives. To quantify the consistency of these human baselines, an Inter-Rater Reliability (IRR) analysis was conducted using the Intraclass Correlation Coefficient (ICC) and Mean Spatial Deviation.

**Table 2**

Inter-Rater Reliability (IRR) Metrics for Ground-Truth Annotation

| Metric | Calculated Value |
| --- | --- |
| Mean Spatial Deviation | 0.1085 km |
| ICC(2,1) Latitude | 1.00 |
| ICC(2,1) Longitude | 1.00 |
| Spatial Tolerance Agreement | 72.72% |

As shown in Table 2, the ICC scores indicate a high degree of absolute agreement between the independent raters. Both latitude and longitude reached 1.00, indicating near-perfect statistical agreement between the independent raters. It is important to note that an ICC of 1.00 does not imply the annotators selected the exact same coordinate pixels; rather, the geographic variance between raters for a single site is microscopic compared to the vast overall spatial distribution of all accidents across the dataset, rendering the inter-rater variance mathematically negligible. This is practically demonstrated by the Mean Spatial Deviation of 0.1085 and a Spatial Tolerance Agreement of 72.72%, which confirm that while identical pins were not placed, human interpretations of the unstructured text remained tightly and reliably clustered. With the solidity of the ground-truth dataset established, we now compare the ALIGN framework's predictions against these verified coordinates.

**Table 3**
Error Metrics across Development, Validation and Verification sets.

| **Error Metric** | **Development Set (N=36)** | **Validation Set (N=77)** | **External Verification by DMP (N=34)** |
|---|---|---|---|
| Mean Error (km) | 0.851 | 0.593 | 0.465 |
| Median Error (km) | 0.358 | 0.265 | 0.334 |
| RMSE (km) | 1.458 | 1.187 | 0.54 |
| **Accuracy Thresholds** | | | |
| Within 0.5 km | 66.70% | 81.82% | 67.60% |
| Within 1.0 km | 77.80% | 89.61% | 94.10% |
| Within 2.0 km | 88.90% | 90.91% | 100.00% |
| Within 5.0 km | 100.00% | 100.00% | 100.00% |

Table 3 summarizes the error metrics across the Development, Validation, and External Verification datasets, demonstrating the ALIGN framework's high precision. The system showed marked improvement from the Development Set to the Validation Set, where the mean error dropped to 0.593 km and the median error to 0.265 km. Crucially, the External Verification set which comprises of official police data, yielded a mean error of 0.465 km, close to the manual validation results. This convergence statistically confirms the reliability of the system's geolocation capabilities.

Furthermore, the External Verification set achieved the lowest Root Mean Square Error (RMSE) of 0.540 km, indicating exceptional stability in urban environments with minimal outliers. High accuracy consistency is evident, with 89.6% of Validation cases and 94.1% of External Verification cases falling within 1.0 km of the true location.

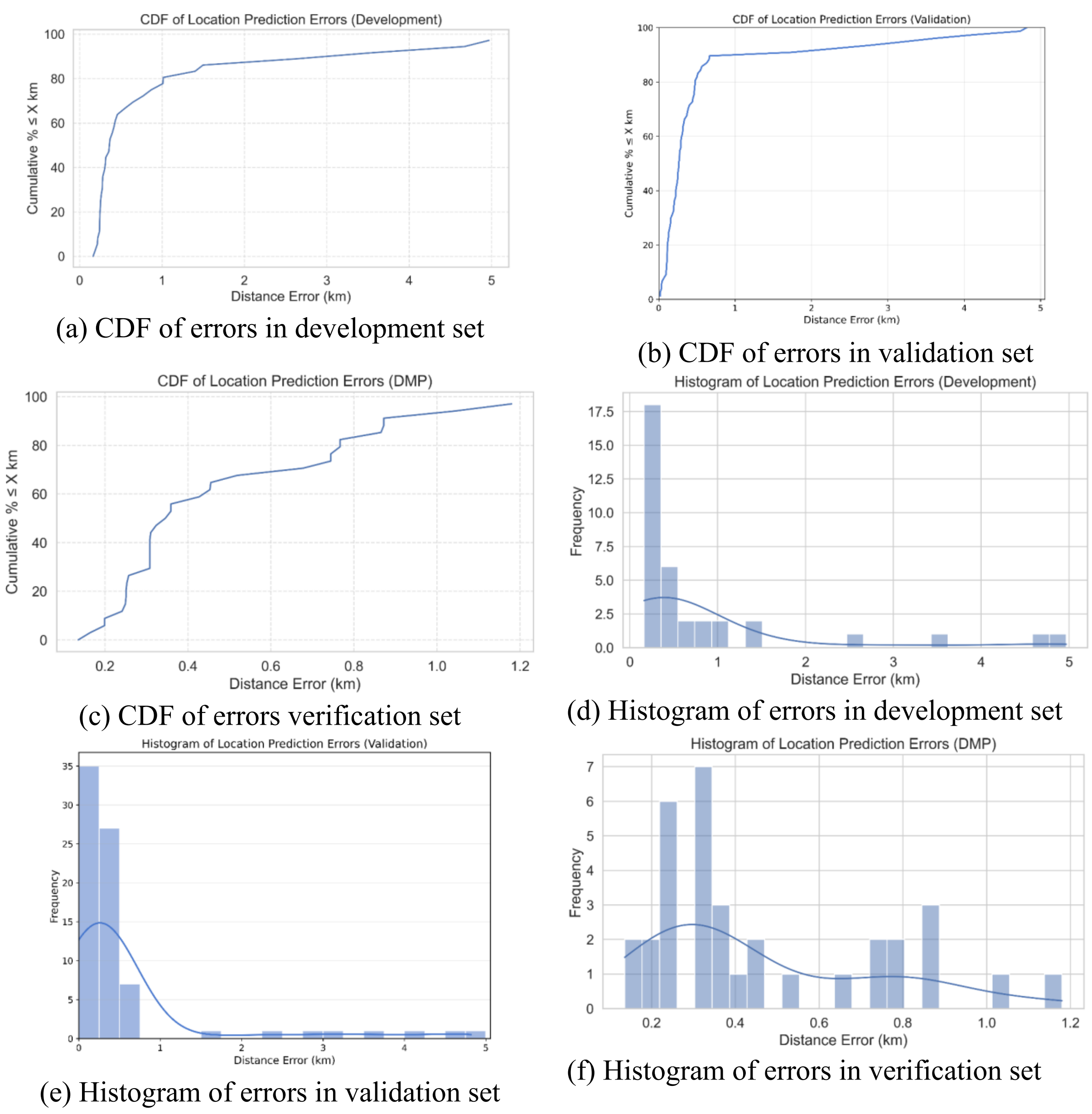

(a) CDF of errors in development set

(b) CDF of errors in validation set

(c) CDF of errors verification set

(d) Histogram of errors in development set

(e) Histogram of errors in validation set

(f) Histogram of errors in verification set

**Fig. 6.** Comparative Error Distributions across Development, Validation, and External Verification sets.

Fig. 6 presents a comparative analysis of error distributions via Cumulative Distribution Functions (CDFs) and histograms. The CDF plots (a–c) illustrate the system's high precision. The development set (a) shows a steep initial rise, with approximately 66.7% of errors under 0.5 km, though the curve indicates the presence of outliers designed to stress-test the system. The validation set (b) exhibits an even steeper ascent, with over 80% of predictions within 500 meters, demonstrating superior generalization on unseen data. Similarly, the external verification set (c) confirms robust performance, with 94.1% of errors falling within 1 km. The histograms (d–f) further clarify these trends. The development histogram (d) displays a strong right skew but includes visible outliers in the 2–5 km range. In contrast, the validation histogram (e) reveals a tighter concentration of errors near zero, confirming the reduction of extreme errors. Finally, the verification histogram (f) shows a distinct clustering of errors between 0.2 km and 0.4 km.

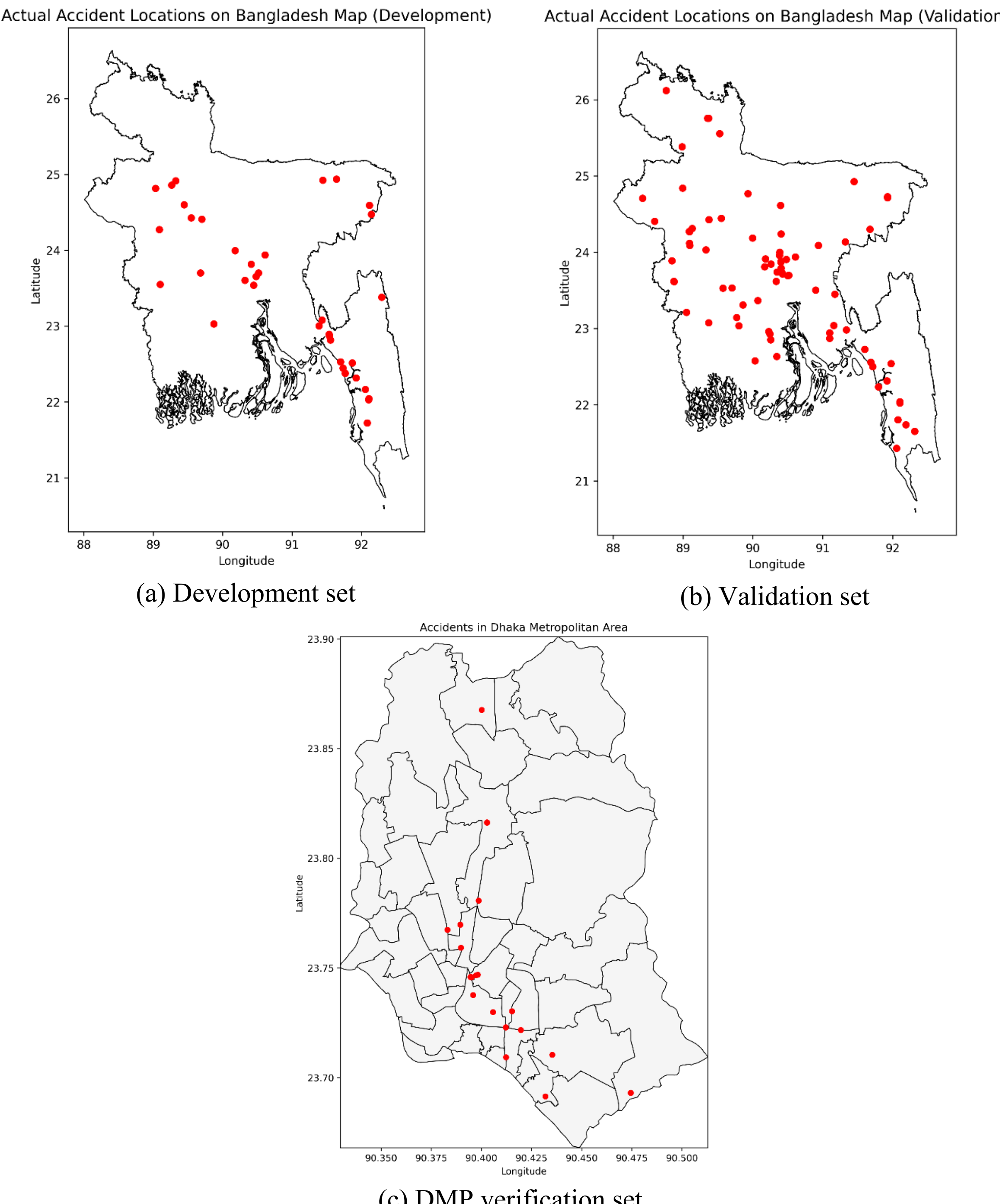

(a) Development set

(b) Validation set

(c) DMP verification set

**Fig. 7.** Spatial distribution of ground-truth locations for (a) Development Set, (b) Validation Set, and (c) DMP Verification Set.

Fig. 7 illustrates the spatial coverage of the ground-truth datasets. Maps (a) and (b) demonstrate that the Development and Validation sets span the entire geography of Bangladesh, capturing diverse environments from northern hills to southern coastal regions. The Validation set (b) displays a denser, more representative national distribution compared to the smaller Development set (a). In contrast, map (c) visualizes the DMP External Verification set, which is strictly concentrated within the Dhaka Metropolitan Area. This comparison confirms that the system was tested across a robust variety of spatial contexts, ranging from broad rural districts to complex, high-density urban street networks.

### *4.2 Comparison with Existing Geo-parsing Baselines*

To contextualize the effectiveness of our approach, we benchmarked it against a representative baseline system inspired by Idakwo et al. (2025). Their framework utilizes a domain-specific NER model trained using spaCy and a rule-based rtc_site generation logic, followed by Google Maps geocoding to derive coordinates. Our implementation closely followed this pipeline using a Gemini-based schema-driven entity extractor customized for the Bangladeshi context, paired with a similar rule-based location composition and Google Geocoding.

We applied this "Text + Geocoding only" pipeline to our validation set. The performance comparison is presented in Table 4.

**Table 4**

Performance comparison against baseline

| Metric | Text + Geocoding Baseline | Proposed System (Validation) |
| --- | --- | --- |
| Mean Error (km) | 10.915 | 0.593 |
| Median Error (km) | 1.233 | 0.265 |
| RMSE (km) | 26.306 | 1.187 |
| Within 0.5 km (%) | 37.66% | 81.82% |
| Within 1 km (%) | 44.16% | 89.61% |
| Within 2 km (%) | 54.55% | 90.91% |
| Within 5 km (%) | 64.94% | 100.0% |

As Table 4 clearly shows, our proposed system outperforms the baseline significantly across all error metrics. Specifically, mean error is reduced from 10.915 km to 0.593 km, a 94.5% reduction. In addition, RMSE, which penalizes large deviations more strongly, drops dramatically from 26.306 km to 1.187 km. Furthermore, accuracy within close thresholds improves substantially. For instance, the proportion of predictions within 1 km jumps from 44.16% to 89.61%.

**Table 5**
Qualitative comparison against existing geolocation frameworks

| Capability | CLAVIN / Mordecai | LNEx | GeoGPT / DEES | RTC-NER | ALIGN (Proposed) |
|---|---|---|---|---|---|
| Neural/Contextual Reasoning | ~ (Limited) | × | ✓ | × | ✓ (Multimodal) |
| Visual Map & OCR Verification | × | × | × | × | ✓ |
| Multilingual (Bangla + English) | × | × | × | × | ✓ |
| Multistage Fallback Logic | × | × | × | × | ✓ |
| Training-free | × | × | × | ✓ | ✓ |
| Fine-Grained Accuracy (<700m) | × | × | Partial | Partial | ✓ |

Table 5 compares ALIGN with existing geolocation frameworks, highlighting its comprehensive multimodal reasoning and engineering efficiency. Unlike prior text-only systems such as CLAVIN, Mordecai, or LNEx, ALIGN uniquely integrates vision–language map verification, OCR-based label recognition, and a multistage fallback logic for enhanced robustness. Moreover, it operates without training, requiring only lightweight calibration—eliminating costly dataset preparation and GPU resources. These innovations enable high-accuracy (~500 m) coordinate localization and full automation for Bangla and English accident reports; positioning ALIGN as the first low-resource multimodal GeoAI framework for real-time accident mapping and safety analytics.

These results underscore that our system has achieved superior performance compared to baseline in this task domain. To the best of our knowledge, no prior geo-parsing system for traffic incidents—particularly in low-resource settings—has demonstrated this level of geographic precision.

*4.3 Ablation Study*

To rigorously quantify the architectural contribution of each module within the ALIGN pipeline, a comprehensive ablation study was conducted. The evaluation utilized the benchmark dataset of 77 Bangla news articles describing road crashes. By systematically disabling core framework components, such as deterministic OCR, road network metadata injection, and the geometric spatial voting loop, we assessed their direct impact on localization accuracy, computational cost, and processing time. It is important to note that all cost calculations reported in this study are based on the official developer API pricing rates active as of March 2026. This quantitative approach validates the necessity of our defense-in-depth mitigation strategy against standard single-pass multimodal inferences. The comparative performance metrics across all structural configurations are detailed in Table 6.

**Table 6**

Ablation Study Table

| Configuration | Successful Entries (N) | Mean Error (km) | Median Error (km) | Accuracy @ 1km (%) | Accuracy @ 2km (%) | Average Tokens per Article | AI Inference Time(s) per Article | Web/ Selenium Time(s) per Article | OCR Processing Time (s) per Article | Average Cost Per Article ($) |
|---|---|---|---|---|---|---|---|---|---|---|
| Stage 1 (Simple Geocoding) | 77/77 | 10.915 | 1.233 | 44.2 | 54.5 | 1,068 | --- | --- | --- | 0.00017 |
| Stage 2 (Grid Scan) | 45/77 | 0.654 | 0.291 | 84.4 | 93.3 | 25,624 | 87.9 | 46.9 | 269.4 | 0.00590 |
| Stage 3 (Fallback) | 32/77 | 1.060 | 0.431 | 65.6 | 81.2 | 130,872 | 369 | 92.1 | 906.4 | 0.02971 |
| w/o OCR | 75/77 | 24.959 | 0.366 | 69.3 | 74.7 | 38,724 | 122.1 | 42.9 | --- | 0.00958 |
| w/o Road Injection | 69/77 | 15.991 | 0.470 | 65.2 | 72.5 | 79,650 | 238.3 | 84.9 | 831.3 | 0.01922 |
| w/o Geometric Voting | 71/77 | 12.707 | 0.462 | 63.4 | 73.2 | 76,372 | 224.6 | 73.3 | 587.1 | 0.01746 |
| Full ALIGN Pipeline | 77/77 | 0.593 | 0.265 | 89.61 | 90.91 | 69,363 | 204.7 | 65.7 | 534.1 | 0.01579 |

Beyond spatial accuracy, the ablation study highlights the operational trade-offs of the framework. The integration of deterministic Optical Character Recognition (OCR) is crucial for accurate multimodal reasoning. When OCR is disabled, the mean error drastically increases from 0.593 km to 24.959 km, and 1 km accuracy drops from 89.61% to 69.3%. This confirms that supplying the Vision-Language Model with explicit text extracted from map tiles effectively prevents visual hallucinations. However, this precision comes at a computational cost. An analysis of system latency reveals that OCR processing is consistently the most time-consuming component of the pipeline, averaging 534.1 seconds per article in the full configuration, compared to 204.7 seconds for AI inference and 65.7 seconds for web automation. Furthermore, token consumption varies significantly based on the required reasoning depth. While the full pipeline averages 69,363 tokens per article, complex cases requiring Stage 3 fallback mechanisms demand up to 130,872 tokens. Ultimately, the system design accepts higher deterministic processing times and token usage to guarantee the sub-kilometer accuracy required for reliable road safety analytics, filtering out visual hallucinations and reducing mean error from 24.959 km to 0.593 km. Ultimately, the fully integrated ALIGN pipeline achieves a sub-kilometer median localization of 0.265 km, representing an 18-fold improvement over baseline geocoding.

### *4.4 Cross-Model Generalizability and Cost Analysis*

To validate the architectural robustness of the ALIGN framework beyond a single proprietary foundation model, a comparative cross-model evaluation was conducted. We benchmarked the system's reasoning backbone using three diverse Large Language Models: Gemini 2.5 Flash, GPT-5-mini, and Llama-4-Maverick. All models were subjected to the exact same 77-entry validation dataset to assess their capability to execute the complex, multi-stage geospatial reasoning tasks required by the pipeline. This analysis evaluated not only localization accuracy and success rates but also computational efficiency, token consumption, and API inference costs, as detailed in Table 7.

**Table 7**
Cross model and cost analysis

| Metric | Gemini 2.5 Flash | GPT-5-mini | Llama-4-Maverick |
|---|---|---|---|
| Successful Entries (N) | 77/77 | 67/77 | 66/77 |
| Mean Error (km) | 0.593 | 4.671 | 9.374 |
| Median Error (km) | 0.265 | 0.895 | 0.562 |
| Acc@1km (%) | 89.61 | 52.24 | 53.03 |
| Acc@2km (%) | 90.91 | 71.64 | 68.18 |
| Average Tokens | 69,363 | 102,194 | 45,127 |
| AI Inference Time (s) | 204.7 | 385.4 | 75.9 |
| Average Cost per Article ($) | 0.01579 | 0.07100 | 0.00226 |

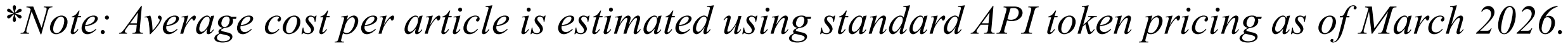

**Note: Average cost per article is estimated using standard API token pricing as of March 2026.*

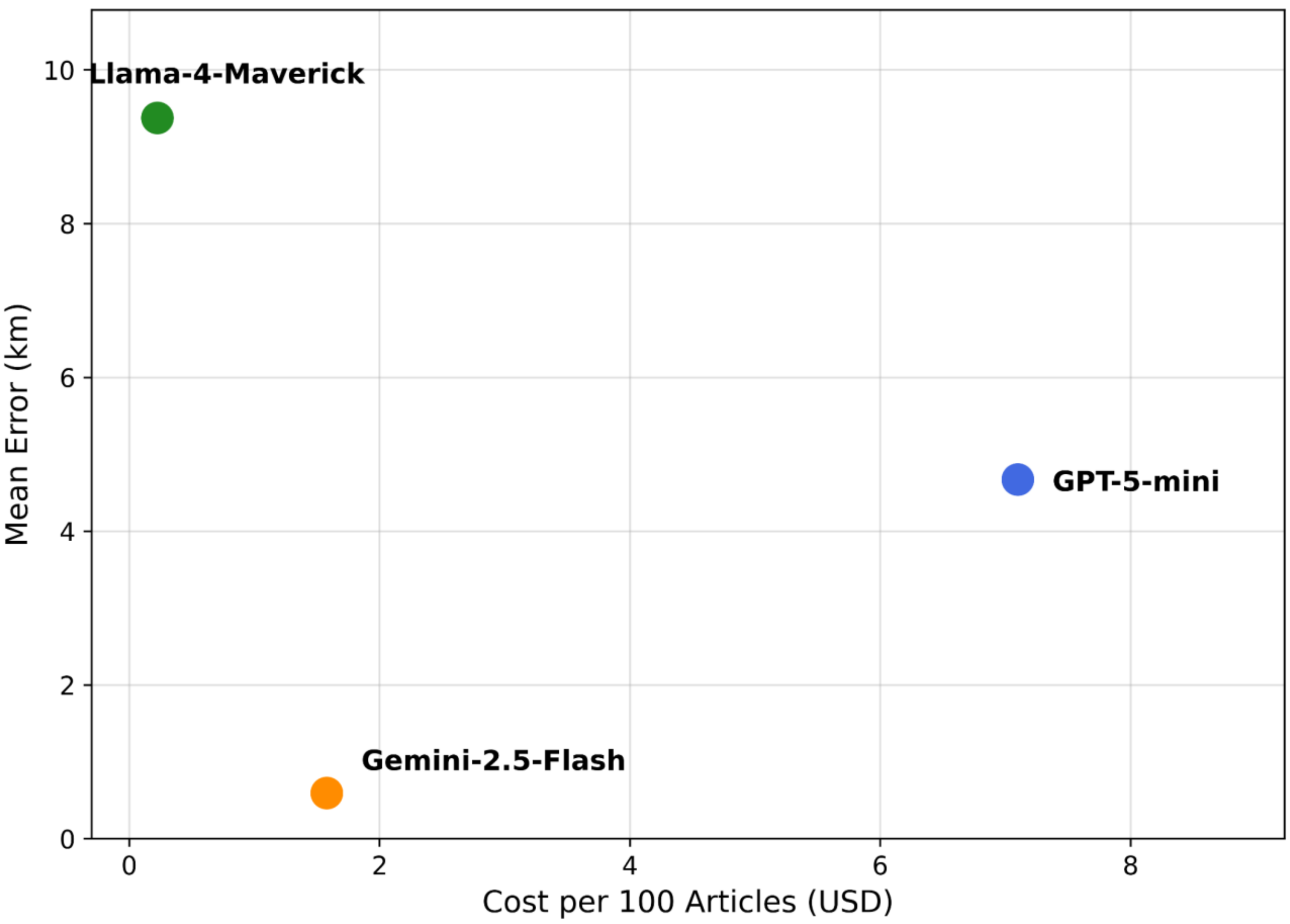

**Fig. 8.** Performance-Cost Pareto front across evaluated reasoning models

To mathematically justify our final model choice for the ALIGN pipeline, we plotted the mean error against estimated token costs for four competitive models (Fig. 8). This analysis reveals a clear Pareto front, illustrating the necessary trade-offs for LMIC deployment. While the open-weights Llama-4-Maverick model represents the most economical option (<$0.010 per article), it also suffers from a catastrophic 9.374 km mean error, which would fail to diagnosably locate black spots. Conversely, while GPT-5-mini offered improved performance, its costs remained high. Ultimately, Gemini-2.5-Flash was scientifically selected as the optimal choice. As Fig. 8 confirms, it provides the mandatory sub-kilometer precision (~0.6 km) necessary for meaningful road safety analysis, while remaining financially viable ($0.016 per article) and competitive with much larger models.

*4.5 Discussion*

The superior performance observed in the validation set compared to the development set reflects the intentional inclusion of "stress-test" cases in the latter to refine fallback logic. Crucially, external verification against official Dhaka Metropolitan Police records yielded a mean error of 0.465 km, which is close to the manual validation set's 0.593 km. This statistical convergence objectively validates the manual annotation methodology used for ground truth and confirms the system's reliability. Furthermore, the system demonstrated enhanced precision in dense urban environments (RMSE 0.540 km), where distinct landmarks help eliminate outliers compared to the national average.

Despite these strengths, several limitations remain. The study focused exclusively on Bangladesh due to the linguistic familiarity required for manual validation, and the primary validation coordinates were derived via Google Maps due to the critical shortage of official national crash data. While the defense-in-depth strategy and 3-run geometric voting effectively reduced severe hallucinations, occasional plausible but incorrect inferences persist in highly ambiguous edge cases. Additionally, as demonstrated in our cross-model evaluation, achieving sub-kilometer precision currently necessitates reliance on proprietary VLM APIs; while open-weights models (such as Llama-4-Maverick) are highly scalable, they presently struggle to maintain the high-fidelity spatial reasoning required for this specific multimodal task.

Regarding transferability, ALIGN's modular architecture is well-suited for other low-resource regions, such as India or Myanmar, which share similar administrative hierarchies and transliteration challenges. Deploying the framework in these areas would primarily require updates to alias databases and prompt schemas. In contrast, adapting the system for high-income regions like the United States or Europe would necessitate structural modifications to handle different spatial units (e.g., counties) and distinct reporting styles. Ultimately, ALIGN's primary value lies in its ability to provide cost-effective, automated geospatial intelligence for data-sparse environments that currently lack robust monitoring infrastructure.

While the deterministic chunk-wise OCR pre-filtering is vital for eliminating Type B visual hallucinations, it introduces a significant latency bottleneck, currently requiring approximately 13 minutes per article. Future iterations of the ALIGN framework can substantially improve processing efficiency by directly addressing this OCR bottleneck. Specifically, optimization efforts will focus on four key areas: (1) deploying more powerful GPU hardware to accelerate baseline inference speeds; (2) parallelizing the chunk-wise OCR processing tasks so multiple map segments can be evaluated simultaneously; (3) developing a custom, lightweight OCR architecture specifically tuned for low-resolution Bangla map labels; and ultimately (4) fine-tuning the foundational Vision-Language Models directly on curated map-label datasets to natively enhance their internal text-recognition capabilities. This final approach aims to completely omit the external OCR dependency, integrating extraction and visual verification into a single, high-speed multimodal inference step.

## 5. Conclusion

This research presented ALIGN, a novel multimodal, multi-stage framework that automates the human-expert process of geolocating road traffic crashes from unstructured news articles. By synergizing LLM-based textual intelligence with VLM-driven visual map verification and OCR, ALIGN successfully resolves the ambiguities that plague traditional text-only systems. Our validation in Bangladesh demonstrated useful performance, achieving a 0.593 km mean error, a 94.5% reduction over baselines. This work delivers an effective methodology for generating the high-precision "black spot" data essential for road safety policy in data-sparse LMICs.

While ALIGN marks a significant advancement, future work will focus on increasing processing efficiency, specifically by addressing the OCR latency bottleneck through hardware acceleration, task parallelization, and the development of a lightweight, domain-specific OCR module. We will also work toward a more readily transferable system design to reduce the domain-specific tuning required for new regions. Finally, further exploration into deeper Agentic AI integration promises to enhance the system's autonomy and reasoning capabilities, moving closer to a fully automated geospatial intelligence agent.

**CRediT authorship contribution statement**

**MD Thamed Bin Zaman Chowdhury:** Conceptualization, Methodology, Software, Validation, Formal analysis, Investigation, Data Curation, Writing – original draft, Visualization. **Moazzem Hossain:** Methodology, Validation, Writing – review & editing, Supervision, Project administration.

**Declaration of competing interest**

The authors declare that they have no known competing financial interests or personal relationships that could have appeared to influence the work reported in this paper.

**Acknowledgments**

This research did not receive any specific grant from funding agencies in the public, commercial, or not-for-profit sectors. We thank Dhaka Metropolitan Police (DMP) for assisting us by providing valuable data.

**Data availability**

The complete source code and the validation dataset generated during this study are publicly available in the project's GitHub repository at https://github.com/Thamed-Chowdhury/ALIGN. To ensure long-term accessibility and reproducibility, the repository has also been archived on Zenodo (DOI: 10.5281/zenodo.18903030). More detailed or extended data can be provided upon reasonable request to the corresponding authors.